\documentclass[10pt,twocolumn,letterpaper]{article}

\usepackage{cvpr}
\definecolor{cvprblue}{rgb}{0.21,0.49,0.74}
\usepackage[pagebackref,breaklinks,colorlinks,allcolors=cvprblue]{hyperref}

\usepackage{multirow}
\usepackage{makecell}
\usepackage{graphicx}
\def\paperID{40897} 
\def\confName{CVPR}
\def\confYear{2026}

\title{
Catalyst4D: High-Fidelity 3D-to-4D Scene Editing via Dynamic Propagation
}

\author{
    Shifeng Chen\textsuperscript{\rm 1,2},
    Yihui Li\textsuperscript{\rm 1,2},
    Jun Liao\textsuperscript{\rm 3},
    Hongyu Yang\textsuperscript{\rm 3}\thanks{Corresponding author},
    Di Huang\textsuperscript{\rm 1,2,4} \\
    \textsuperscript{\rm 1} State Key Laboratory of Complex and Critical Software Environment, Beihang University, China \\
    \textsuperscript{\rm 2} School of Computer Science and Engineering, Beihang University, China\\
    \textsuperscript{\rm 3} School of Artificial Intelligence, Beihang University, China\\
    \textsuperscript{\rm 4} Zhejiang Industrial Big Data and Robot Intelligent System Key Laboratory,\\ Hangzhou Innovation Institute, Beihang University, China\\
    {\tt\small \{shifengchen, kidleyh, junliao, hongyuyang, dhuang\}@buaa.edu.cn}
}
\begin{document}
\twocolumn[{
\renewcommand\twocolumn[1][]{#1}
\maketitle
\begin{center}
    \centering
    \captionsetup{type=figure}
    \includegraphics[width=\textwidth]{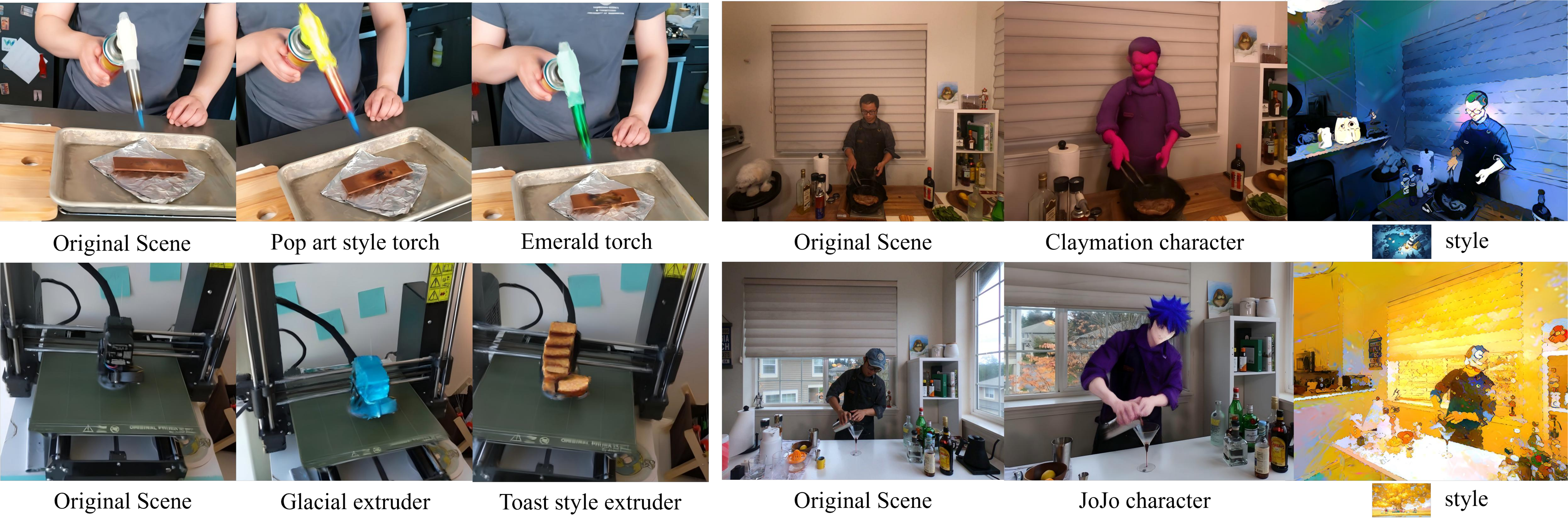}
    \captionof{figure}{We present \textbf{Catalyst4D}, a framework that propagates single-frame 3D edits to dynamic sequences. It excels at both precise local modifications and high-quality global style transfer. Catalyst4D demonstrates robust performance on both monocular (left) and multi-camera (right) scenes. Please refer to the supplementary material for more intuitive visual results.}
    \label{fig:results}
\end{center}
}]
{
  \renewcommand{\thefootnote}{\fnsymbol{footnote}}
  \footnotetext[1]{Corresponding author}
}
\begin{abstract}
Recent advances in 3D scene editing using NeRF and 3DGS enable high-quality static scene editing. In contrast, dynamic scene editing remains challenging, as methods that directly extend 2D diffusion models to 4D often produce motion artifacts, temporal flickering, and inconsistent style propagation.
We introduce \textbf{Catalyst4D}, a framework that transfers high-quality 3D edits to dynamic 4D Gaussian scenes while maintaining spatial and temporal coherence.  
At its core, \textbf{Anchor-based Motion Guidance (AMG)} builds a set of structurally stable and spatially representative anchors from both original and edited Gaussians. These anchors serve as robust region-level references, and their correspondences are established via optimal transport to enable consistent deformation propagation without cross-region interference or motion drift.  
Complementarily, \textbf{Color Uncertainty-guided Appearance Refinement (CUAR)} preserves temporal appearance consistency by estimating per-Gaussian color uncertainty and selectively refining regions prone to occlusion-induced artifacts.  
Extensive experiments demonstrate that Catalyst4D achieves temporally stable, high-fidelity dynamic scene editing and outperforms existing methods in both visual quality and motion coherence. 
\end{abstract}    
\section{Introduction}
Recent advances in 3D scene representation, particularly 3D Gaussian Splatting (3DGS)~\cite{kerbl20233d}, have significantly enhanced the quality and flexibility of static scene editing. Building upon this representation, recent studies~\cite{chen2024dge,kim2024dreamcatalyst,galerne2025sgsst,wen2025intergsedit,Zhao_2025_ICCV,wynn2025morpheus} have enabled high-quality and versatile 3D editing, supporting fine-grained object manipulation and global style transfer with strong spatial consistency. These developments demonstrate the expressive power of 3DGS in jointly capturing geometry and appearance, providing a solid foundation for controllable and photorealistic scene editing in static environments.

In contrast, editing dynamic 4D scenes remains a far more challenging and underexplored problem. Although notable progress has been achieved in 4D scene reconstruction~\cite{song2023nerfplayer,xu2024grid4d,lu2024dn,kwak2025modec,li2025time,chen2025dash}, extending editing capabilities to the dynamic domain introduces additional difficulties in maintaining both spatial precision and temporal stability. Existing dynamic editing approaches~\cite{mou2024instruct,he2025ctrl,kwon2025efficient} mainly adapt 2D diffusion models to spatio-temporal settings by fitting 4D representations to edited 2D frames. However, since 2D image edits operate purely in pixel space without explicit geometric reasoning, they often lead to spatial distortions, temporal flickering, and unintended appearance changes in unedited regions. Moreover, instruction-based editors provide ambiguous guidance for localized geometry modification and struggle to propagate artistic style consistently across viewpoints. These limitations highlight the need for geometry-aware editing frameworks that can transfer the strengths of static 3D editing into the dynamic 4D domain.

Recent 4D Gaussian scene models typically represent motion by coupling canonical 3D Gaussians with a learned deformation network~\cite{wuswift4d,wu20244d,xu2024grid4d,chen2025dash,li2025time}, ensuring geometric continuity and semantic correspondence across frames. This representation naturally offers a foundation for extending static 3D edits into dynamic sequences. Yet, reusing the original deformation field after editing often results in severe motion artifacts, particularly when the geometry or appearance of Gaussians is altered through cloning, splitting, or pruning. Because the deformation network is trained exclusively on the original geometry, it lacks motion priors for newly introduced Gaussians and fails to generalize to edited configurations. This fundamentally restricts the applicability of existing 4D representations to dynamic scene editing.

To address these challenges, we present \textbf{Catalyst4D}, a unified framework that transfers high-quality 3D edits to dynamic 4D scenes while maintaining spatial and temporal coherence. Catalyst4D decouples spatial editing from temporal propagation, allowing the edited scene to retain the flexibility of static 3D editors and the structural consistency required for coherent motion over time.

At the core of Catalyst4D lies \textbf{\textit{Anchor-based Motion Guidance (AMG)}}, which provides reliable motion supervision for edited Gaussians. Instead of relying on unstable point-wise correspondences, AMG extracts a sparse set of spatially representative anchors from both the original and edited Gaussian point clouds in the first frame, capturing the underlying object structure. Region-level correspondences are established through optimal transport, enabling localized and semantically consistent deformation while preventing cross-region interference and motion drift.

Complementary to geometry alignment, \textbf{\textit{Color Uncertainty-guided Appearance Refinement (CUAR)}} focuses on temporal appearance consistency. Many Gaussians lack direct color supervision after editing, especially those occluded in early frames, resulting in flickering or inconsistent textures. 
CUAR quantifies the temporal reliability of appearance updates and selectively refines regions with high uncertainty. By propagating confident color information along motion trajectories, Catalyst4D automatically corrects occlusion-induced appearance artifacts and achieves temporally stable rendering.

In summary, Catalyst4D establishes a unified pathway from static 3D editing to dynamic scene manipulation, combining spatial precision with temporal coherence. Our key contributions include:
\begin{itemize}
\item We present \textbf{Catalyst4D}, a unified framework that transfers 3D Gaussian edits to dynamic 4D scenes, achieving high geometric and temporal consistency.
\item We propose \textbf{Anchor-based Motion Guidance (AMG)}, which establishes semantically aligned region-level correspondences to enable localized, coherent deformation and prevent motion drift.
\item We introduce \textbf{Color Uncertainty-guided Appearance Refinement (CUAR)}, which quantifies temporal reliability and refines uncertain regions for stable appearance propagation.
\end{itemize}
Extensive experiments show that Catalyst4D delivers temporally consistent, high-fidelity dynamic scene editing.
\section{Related Work}
\noindent\textbf{Static Scene Editing.}
Neural Radiance Fields (NeRF)~\cite{mildenhall2020nerf} have revolutionized 3D scene reconstruction and inspired numerous editing approaches, including template-based methods using point clouds~\cite{chen2023neuraleditor} or meshes~\cite{yang2022neumesh,yuan2022nerf,jambon2023nerfshop}, as well as text-guided editing~\cite{wang2022clip,wang2023nerf,haque2023instruct,wang2024proteusnerf}. However, the volumetric formulation of NeRFs causes high rendering cost and blurred geometry, limiting efficiency and fidelity.  



The introduction of 3DGS \cite{kerbl20233d, li2025micro,li2025microenhanced, li2026tokensplat, lv2026gau} has addressed many of these limitations by enabling point-based scene representations with efficient rendering and flexible editing. Methods such as GaussiansEditors \cite{chen2024gaussianeditor,wang2024gaussianeditor} constrain gradient-based optimization to target 3D Gaussians, achieving efficient object-level editing while leveraging 2D diffusion models as priors. Subsequent works \cite{chen2024dge,wu2024gaussctrl,wang2024view,wen2025intergsedit} incorporate 3D structural information into diffusion-based editing to improve cross-view consistency, and others \cite{NEURIPS2024_571dd493,wynn2025morpheus,erkocc2025preditor3d,you2025instainpaint} exploit depth information from 3DGS for geometry modification. High-quality stylization methods \cite{liu2024stylegaussian,galerne2025sgsst,Zhao_2025_ICCV} extend image style transfer techniques to 3D scenes, further broadening 3DGS applications in artistic creation. Despite their capabilities, these approaches remain confined to static scenes. Building upon these strong 3D editing foundations, extending them to dynamic scenes with temporal coherence remains an open and compelling challenge.

\begin{figure*}[t]
\centering
\includegraphics[width=\textwidth]{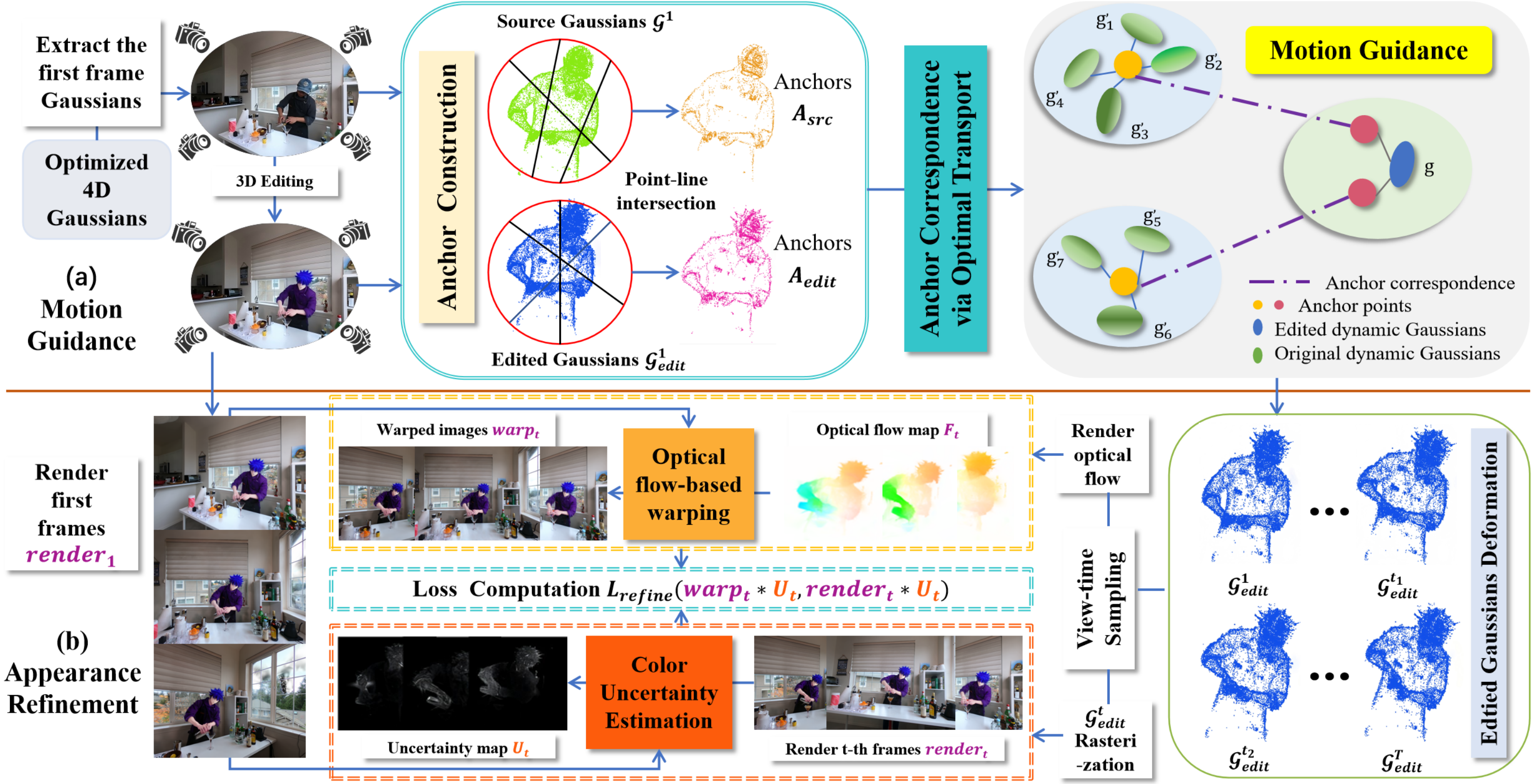}
\caption{Overview of Catalyst4D. Given the first-frame edited dynamic Gaussians, our (a) \textbf{Anchor-based Motion Guidance} establishes region-level correspondences with the original Gaussians via anchor construction and optimal transport, enabling reliable deformation transfer. Then, (b) \textbf{Color Uncertainty-guided Appearance Refinement} leverages first-frame warping and Gaussian color consistency to identify and correct motion-induced artifacts across time.}
\label{fig:pipeline}
\end{figure*}

\noindent\textbf{Dynamic Scene Editing.}
While static scene editing has achieved remarkable progress, extending it to dynamic scenes introduces additional challenges in maintaining temporal and spatial coherence.
Existing approaches \cite{mou2024instruct,he2025ctrl,kwon2025efficient,liu20254dgs,shi2025mono4deditor} generally adapt pre-trained 2D diffusion models \cite{brooks2023instructpix2pix,zhang2023adding} to 4D representations, aligning 4D scenes with 2D supervision.
For instance, Control4D \cite{shao2024control4d} employs a GAN-based discriminator to guide optimization of its Gaussian-Planes representation, while Instruct 4D-to-4D \cite{mou2024instruct} edits individual videos to supervise full 4D optimization.
CTRL-D \cite{he2025ctrl} enhances user control through DreamBooth-style \cite{ruiz2023dreambooth} fine-tuning on a single reference image, and Instruct-4DGS \cite{kwon2025efficient} improves efficiency by separately editing the static component of 4D Gaussian Splatting (4DGS) \cite{wu20244d} and refining it with Score Distillation Sampling (SDS) \cite{poole2022dreamfusion}.
Despite these advances, reliance on 2D guidance often causes degraded spatial-temporal fidelity due to limited 3D consistency. Catalyst4D addresses this issue by extending high-quality 3D editing pipelines to dynamic 4D scenes, ensuring coherent motion and temporally consistent appearance through explicit 3D-to-4D propagation.
\section{Scene Representation}
Our framework builds upon 4D representations that combine static 3D Gaussians with dynamic deformation fields.

\noindent\textbf{3D Gaussian Splatting.} 
We adopt anisotropic 3D Gaussians~\cite{kerbl20233d} to represent the scene. Each Gaussian is defined by a center $\mu \in \mathbb{R}^3$ and a covariance matrix $\Sigma \in \mathbb{R}^{3\times3}$, whose spatial influence at a point $\mathbf{x}$ is given by:
\begin{equation}
G(\mathbf{x}) = \exp\!\left(-\tfrac{1}{2}(\mathbf{x}-\mu)^\mathrm{T}{\Sigma}^{-1}(\mathbf{x}-{\mu})\right).
\end{equation}
The covariance is factorized as ${\Sigma} = \mathbf{R}\mathbf{S}\mathbf{S}^\mathrm{T}\mathbf{R}^\mathrm{T}$, where $\mathbf{R}$ is a rotation matrix (parameterized by quaternion $\mathbf{q}$) and $\mathbf{S}$ is a scaling matrix (from scale vector $\mathbf{s}$).  
Each Gaussian additionally has an opacity $\sigma$ and spherical harmonics coefficients $\mathbf{sh}$ to encode view-dependent color. The full scene can thus be represented as a set:
$\mathcal{G} = \{G_i: {\mu}_i, \mathbf{q}_i, \mathbf{s}_i, \sigma_i, \mathbf{sh}_i\}_{i=1}^N$.
During differentiable rendering, Gaussians are projected to 2D and composited via $\alpha$-blending. The pixel color $C$ along ray $\mathbf{r}$ is:
\begin{equation}
\label{eq:render}
C(\mathbf{r}) = \sum_{i \in \mathcal{N}} \operatorname{SH}(\mathbf{sh}_i, \mathbf{v}) \alpha_i~\!\!\!\prod_{j=1}^{i-1}\!\!~(1-\alpha_j),
\end{equation}
where $\mathcal{N}$ is the depth-sorted list of intersecting Gaussians, $\mathbf{v}$ is the viewing direction, and $\alpha_i$ combines per-Gaussian opacity and projected Gaussian influence.

\noindent\textbf{Dynamic 4D Extension.} 
For dynamic scenes, recent methods~\cite{wuswift4d,wu20244d,xu2024grid4d,li2025time} represent motion by coupling a canonical 3D Gaussian set $\mathcal{G}_c$ with a learned deformation field $\mathcal{F}_{\theta}$ that maps canonical parameters to time $t$:
\begin{equation}
\label{eq:deform}
\Delta {\mu}, \Delta \mathbf{q}, \Delta \mathbf{s} = \mathcal{F}_{\theta}({\mu}_c, t; \mathbf{z}),
\end{equation}
where $\mathbf{z}$ is a scene-level latent code.  
The deformed parameters at time $t$ are computed as:
${\mu}_t = {\mu}_c + \Delta {\mu}$, 
$\mathbf{q}_t = \mathbf{q}_c + \Delta \mathbf{q}$, and 
$\mathbf{s}_t = \mathbf{s}_c + \Delta \mathbf{s}$.  
This deformation-based formulation ensures temporal coherence and semantic consistency across frames.  
Our framework is compatible with various 4D Gaussian formulations, and in practice, we build upon Swift4D~\cite{wuswift4d} and 4DGS~\cite{wu20244d}, which serve as the base representations in our experiments.

\section{Method}
As illustrated in Fig.~\ref{fig:pipeline}, Catalyst4D transfers 3D Gaussian edits from the canonical frame to the entire 4D sequence. 
Given edited first-frame Gaussians obtained by existing 3D Gaussian editors (\textit{e.g.},~\cite{chen2024dge,kim2024dreamcatalyst,galerne2025sgsst}), our goal is to propagate the edits through time while maintaining geometric accuracy and temporal coherence. 
To this end, Catalyst4D introduces two key components:  
\textbf{(1)} \textbf{Anchor-based Motion Guidance (AMG)}, which establishes region-level correspondences via anchors to enable localized, consistent deformation transfer; and  
\textbf{(2)} \textbf{Color Uncertainty-guided Appearance Refinement (CUAR)}, which corrects occlusion-induced color artifacts by propagating confident appearance information along motion trajectories.  

\subsection{Anchor-based Motion Guidance}

Given the canonical 3D Gaussian set $\mathcal{G}_c$ and its deformation field $\mathcal{F}_{\theta}$, 
the first-frame Gaussians can be generally written as:
\begin{equation}
\mathcal{G}^{1} = \mathcal{F}_{\theta}(\mathcal{G}_c, t{=}1).
\end{equation}
After applying edits to this Gaussian representation, we obtain the edited set $\mathcal{G}_{edit}^{\mathrm{1}}$.
The AMG module then transfers these edits across time 
by leveraging semantically aligned anchors.

\noindent\textbf{Anchor Points and Correspondences.}  
To propagate 3D edits reliably across time, it is crucial to establish stable and meaningful correspondences between regions of the original and edited Gaussian clouds. Directly using individual Gaussians is prone to noise and local variations. To address this, we construct region-level anchors on both the original and edited first-frame Gaussian clouds, $\mathcal{G}^{1}$ and $\mathcal{G}_{edit}^{1}$, for motion-guided propagation.

For each Gaussian, we compute $k$-nearest neighbors to form disjoint local neighborhoods $\{\mathcal{N}_{ei}\}$, ensuring coverage while avoiding excessive overlap. Candidate lines are generated by uniformly sampling point pairs on the surface of the minimum bounding sphere $S$ enclosing the dynamic object point cloud. Each point on $S$ is parameterized as:
\begin{equation}
S_r(u,\varphi) = \big(r\sqrt{1-u^2}\cos\varphi,\; r\sqrt{1-u^2}\sin\varphi,\; ru\big),
\end{equation}
with $u\in[-1,1]$, $\varphi\in[0,2\pi)$, and $r$ the sphere radius. A line is considered to intersect a neighborhood $\mathcal{N}_{ei}$ when the entire neighborhood lies within a radius-$\delta$ cylinder centered on that line. For each intersecting neighborhood, a single anchor $\mathbf{p}$ is computed as the distance-weighted centroid:
\begin{equation}
\mathbf{p} \;=\; \frac{\sum_{\mathbf{x}\in\mathcal{N}_{ei}} d_x\,\mathbf{x}}{\sum_{\mathbf{x}\in\mathcal{N}_{ei}} d_x},
\end{equation}
where $d_x$ is the point-to-line distance. Following~\cite{deng2021robust}, we set $\delta = \tfrac{\sqrt{3}}{2}\,d_{\mathrm{mean}}$, where $d_{\mathrm{mean}}$ is the mean intra-neighborhood distance. Only one anchor is retained per neighborhood. 
Applying this procedure to both the original and edited first-frame Gaussian clouds, $\mathcal{G}^{1}$ and $\mathcal{G}_{edit}^{1}$, yields two sets of anchor points: 
\begin{equation}
A_\mathrm{src} = \{\mathbf{p}_i^\mathrm{src}\}_{i=1}^{n}, \quad
A_\mathrm{edit} = \{\mathbf{p}_j^\mathrm{edit}\}_{j=1}^{m},
\end{equation}
where each anchor $\mathbf{p}_i^\mathrm{src}$ or $\mathbf{p}_j^\mathrm{edit}$ represents the 3D position of a structurally representative region in the corresponding Gaussian cloud. 

To establish robust anchor correspondences that account for varying densities and structural changes induced by editing, we compute a soft correspondence matrix 
$\mathbf{P} \in \mathbb{R}^{n \times m}$ using unbalanced optimal transport (UOT)~\cite{burkard1999linear,villani2008optimal}, solved via the Sinkhorn algorithm~\cite{cuturi2013sinkhorn}. 
Here, each element $\mathbf{P}_{ij}$ represents the transport from edited anchor $\mathbf{p}_j^\mathrm{edit}$ to source anchor $\mathbf{p}_i^\mathrm{src}$, and reliable correspondences are written as:
\begin{equation}
corr = \{ (\mathbf{p}_i^\mathrm{src}, \mathbf{p}_j^\mathrm{edit}) \mid i = \operatorname*{arg\,max}_{k} \mathbf{P}_{kj} \}.
\end{equation}
For implementation details, we refer the reader to the supplementary material.

The anchors produced by line-based sampling across the bounding sphere and the neighborhood-based cylinder test are spatially representative and structurally stable, capturing diverse local orientations while filtering out outliers and suppressing local noise. 
Using these anchors together with unbalanced optimal transport, we establish semantically aligned correspondences $corr$ between $A_\mathrm{src}$ and $A_\mathrm{edit}$, providing a stable foundation for propagating 3D edits.

\noindent\textbf{Anchor-Driven Deformation Aggregation.}
This stage aims to propagate the temporal deformations of source Gaussians to the edited Gaussians $\mathcal{G}_\mathrm{edit}^1$.
For each edited Gaussian $\mathbf{g} \in \mathcal{G}_\mathrm{edit}^1$, we first identify its influencing anchors $A^{\mathrm{sub}}_{\mathrm{edit}} \subset A_\mathrm{edit}$, and use the correspondence mapping $corr$ to locate the matched source anchors $A^{\mathrm{sub}}_{\mathrm{src}} \subset A_\mathrm{src}$. We then retrieve the source Gaussians $\mathcal{G}^{\mathrm{1,sub}}_{\mathrm{src}} \subset \mathcal{G}_\mathrm{src}^1$ that contributed to these anchors, and aggregate their temporal deformations to guide the motion of $\mathbf{g}$ over time.

Specifically, at timestamp $t$, the position deformation $\Delta{\mu}_\mathbf{g}^t$ is computed as a weighted average of the corresponding source Gaussians' deformations:
\begin{equation}
    \Delta{\mu}_\mathbf{g}^t = \frac{\sum_{\mathbf{g}' \in \mathcal{G}^{\mathrm{1,sub}}_{\mathrm{src}}} w_{\mathbf{g}'} \Delta{\mu}_{\mathbf{g}'}^t}{\sum_{\mathbf{g}' \in \mathcal{G}^{\mathrm{1,sub}}_{\mathrm{src}}} w_{\mathbf{g}'}} ,
\end{equation}
where $\Delta{\mu}_{\mathbf{g}'}^t$ is the temporal deformation of source Gaussian $\mathbf{g}'$ from frame 1 to $t$, predicted by the deformation network. The weight $w_{\mathbf{g}'}$ combines the Gaussian's opacity $\sigma_{\mathbf{g}'}$ and its Mahalanobis distance to $\mathbf{g}$:
\begin{equation}
    w_{\mathbf{g}'} = \sigma_{\mathbf{g}'} \exp\Big(-\frac{1}{2}({\mu}_{\mathbf{g}'} - {\mu}_\mathbf{g})^\mathrm{T} {\Sigma}_{\mathbf{g}'}^{-1} ({\mu}_{\mathbf{g}'} - {\mu}_\mathbf{g})\Big).
\end{equation}
Rotation $\Delta \mathbf{q}_\mathbf{g}$ and scale deformation $\Delta \mathbf{s}_\mathbf{g}$ are computed analogously. In this way, we can achieve robust and coherent motion while maintaining geometric fidelity. Further implementation details are provided in the supplementary material.

\subsection{Color Uncertainty-guided Appearance Refinement}

Even with anchor-driven motion guidance, edited dynamic Gaussians $\mathcal{G}_{\mathrm{edit}}^1$ can still exhibit color artifacts due to motion propagation or occlusions.  
While recent works rely on diffusion-based image enhancement~\cite{mou2024instruct,he2025ctrl,zhong2025taming}, post hoc corrections may introduce inconsistencies with the original scene appearance.  
In contrast, we exploit the reliability of the edited first-frame, which are generated by 3D editing methods that inherently maintain multi-view consistency.  
These first-frame images are warped to subsequent frames using Gaussian-driven optical flow to supervise and refine the appearance of $\mathcal{G}_{\mathrm{edit}}^1$ throughout the sequence. This process is illustrated in the bottom part of Fig.~\ref{fig:pipeline}.

\noindent\textbf{Optical Flow Rendering.}  
Given the temporal deformation $\Delta{\mu}^t$ from motion guidance, we first render optical flow maps describing the image-space displacement of each Gaussian from frame 1 to frame $t$.  
Inspired by~\cite{kim20244d,zhou2024hugs}, for a viewpoint $v$, the 3D positions ${\mu}_1$ and ${\mu}_t$ are projected to 2D via the projection function $\operatorname{Proj}_v$, yielding the per-Gaussian displacement:
\begin{equation}
    f_{1 \to t}^v = \operatorname{Proj}_v({\mu}_t) - \operatorname{Proj}_v({\mu}_1).
\end{equation}
These displacements are then $\alpha$-blended in 3DGS to obtain the rendered optical flow map:
\begin{equation}
    F_{1 \to t}^v = \sum_{i \in N} f_{i,1 \to t}^v \alpha_i \prod_{j=1}^{i-1}(1-\alpha_j).
\end{equation}
Finally, the frame-1 image can be warped to frame $t$:
\begin{equation}
    \operatorname{warp}_t^v = \operatorname{render}_1^v + F_{1 \to t}^v.
\end{equation}

\noindent\textbf{Color Uncertainty Estimation.}  
To identify regions prone to appearance artifacts, we estimate the color uncertainty of each Gaussian based on temporal appearance inconsistency.  
For view $v$, the projected color difference between frame 1 and frame $t$ is computed as:
\begin{equation}
    C_{\mathrm{diff}}^{v,t} = \Vert\operatorname{SH}(\mathbf{sh},\mathbf{v})_t - \operatorname{SH}(\mathbf{sh},\mathbf{v})_1\Vert_1,
\end{equation}
and the Gaussian-wise color uncertainty is defined as:
\begin{equation}
    \xi_t^v = 1 - \exp(-C_{\mathrm{diff}}^{v,t}).
\end{equation}
Larger values of $\xi_t^v$ indicate higher likelihood of color artifacts in frame $t$ for view $v$.  
These uncertainties are composited into a per-pixel uncertainty map $U_t^v$ using $\alpha$-blending:
\begin{equation}
    U_t^v = \sum_{i \in \mathcal{N}} \xi_{i,t}^v \alpha_i \prod_{j=1}^{i-1}(1-\alpha_j),
\end{equation}
and subsequently binarized to produce an artifact mask:
\begin{equation}
    M_t^v = \big(U_t^v > \epsilon \cdot \operatorname{mean}(U_t^v)\big),
\end{equation}
where $\operatorname{mean}(U_t^v)$ is the average of the uncertainty map, and $\epsilon$ is a hyperparameter controlling the mask coverage.

\begin{figure*}[t]
\centering
\includegraphics[width=\textwidth]{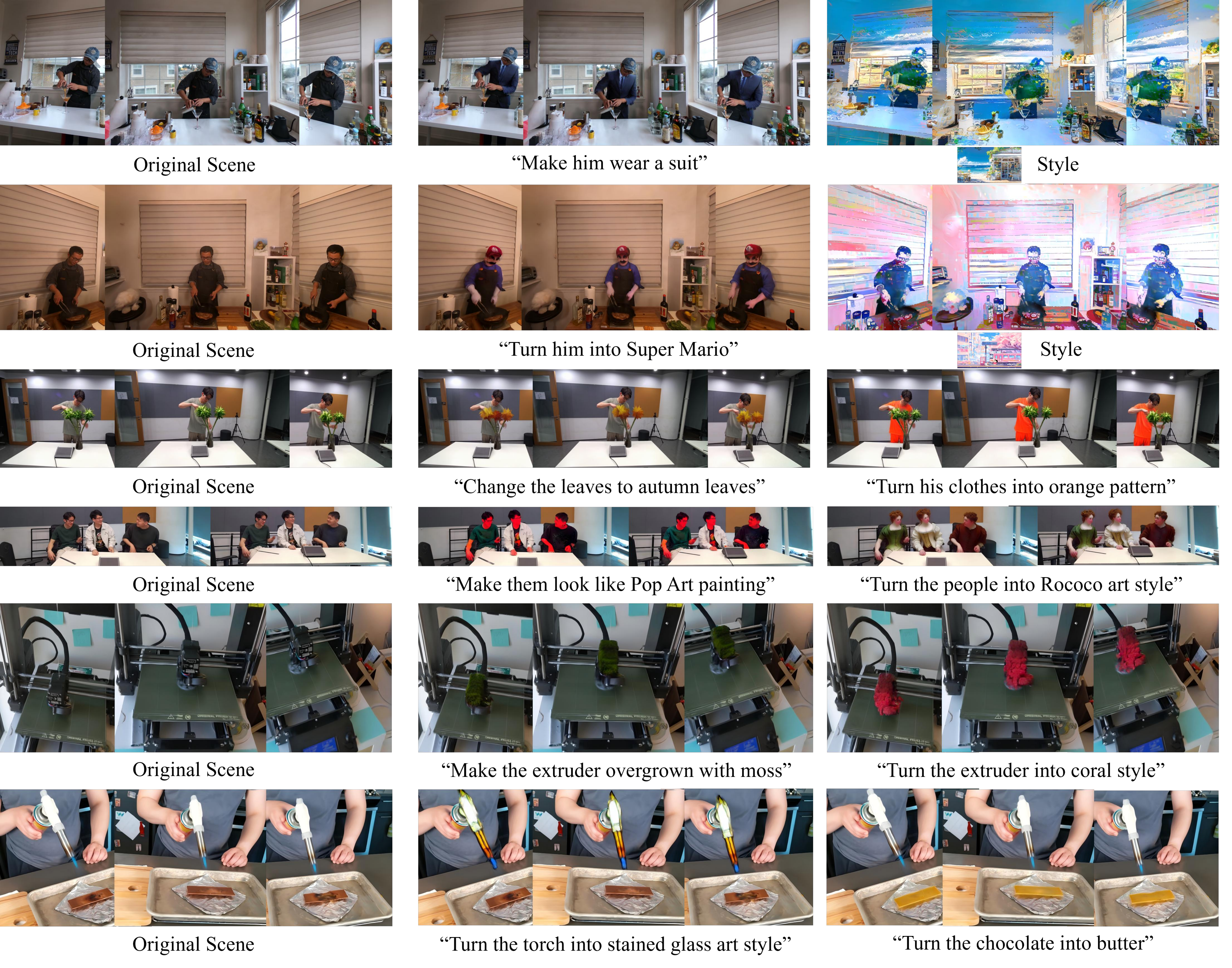}
\caption{Qualitative editing results across multiple scenes: \textit{Cut-beef}, \textit{Coffee-martini} and \textit{Sear-steak} (DyNeRF), \textit{Discussion} and \textit{Trimming} (MeetRoom), \textit{3Dprinter} and \textit{Torchocolate} (HyperNerf). Our method successfully edits dynamic scenes while adhering to user instructions.}
\label{fig:ours_result}
\end{figure*}

\begin{figure*}[t]
\centering
\includegraphics[width=\textwidth]{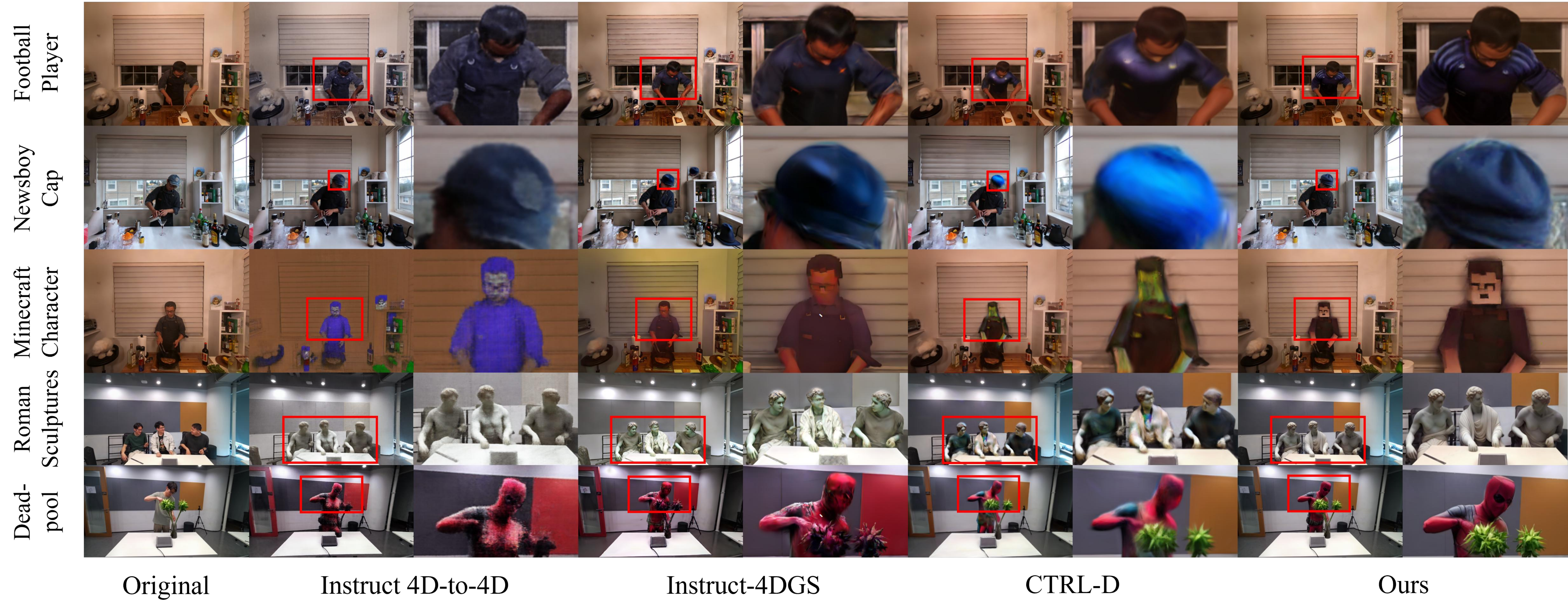}
\caption{Qualitative comparison with Instruct 4D-to-4D, Instruct-4DGS and CTRL-D. Red boxes indicate magnified regions. While competing methods often cause unintended modifications to non-target regions, Catalyst4D demonstrates precise, localized editing.}
\label{fig:qualitative}
\end{figure*}

\noindent\textbf{Refinement Loss.} 
To correct the artifact regions identified by $M_t^v$, we use the warped image $\operatorname{warp}_t^v$ as pseudo ground-truth supervision.  
We define a foreground refinement loss over the artifact mask:
\begin{equation}
\begin{split}
    L_{\mathrm{fore}} = (1 - \eta)\Vert M_t^v \odot (\operatorname{render}_t^v - \operatorname{warp}_t^v)\Vert_1 \\
    + \eta L_{\mathrm{ssim}}(M_t^v \odot \operatorname{render}_t^v, M_t^v \odot \operatorname{warp}_t^v)
\end{split}
\end{equation}
where $\odot$ denotes the Hadamard product, and $L_{\mathrm{ssim}}$ promotes perceptual similarity. Meanwhile, to preserve non-artifact regions, we impose a background regularization loss:
\begin{equation}
    L_{\mathrm{back}} = \Vert(1 - M_t^v) \odot (\operatorname{render}_t^v - \operatorname{render}_{t,\mathrm{org}}^v)\Vert_1,
\end{equation}
Here, $\mathrm{render}_{t,\mathrm{org}}^v$  is the original rendering before this refinement step. The final objective combines the two:
\begin{equation}
    L_{\mathrm{refine}} = (1 - \zeta)L_{\mathrm{fore}} + \zeta L_{\mathrm{back}}.
\end{equation}

By incorporating a color uncertainty-driven mask to selectively refine artifact regions while preserving uncorrupted areas, our method achieves localized appearance correction without introducing global inconsistencies.
\section{Experiment}
\subsection{Experimental Setup}
\noindent\textbf{Dataset and Metrics.} 
We conduct experiments on the multi-camera video datasets DyNeRF \cite{li2022neural} and MeetRoom \cite{li2022streaming}, as well as the monocular dataset HyperNeRF \cite{park2021hypernerf}.  We compare our method with SOTA approaches: Instruct 4D-to-4D \cite{mou2024instruct}, CTRL-D \cite{he2025ctrl}, and Instruct-4DGS \cite{kwon2025efficient}.
The results of all methods are evaluated on test views using CLIP \cite{radford2021learning} similarity and a temporal consistency score \cite{huang2024vbench}.

\noindent\textbf{Implementation Details.} 
We incorporate several baseline 3D editing methods, including the 3D-consistent editing approach DGE \cite{chen2024dge}, the SDS inversion-based framework DreamCatalyst \cite{kim2024dreamcatalyst}, and the high-resolution style transfer method SGSST \cite{galerne2025sgsst}. For 4D scene representation, we employ Swift4D \cite{wuswift4d} under multi-camera settings, while utilizing 4DGS \cite{wu20244d} for monocular scenes. 

\subsection{Qualitative Results}
\noindent\textbf{Visual Examples of Catalyst4D.} 
Qualitative results on three datasets are presented in Fig.~\ref{fig:results} and Fig.~\ref{fig:ours_result}. 
Built upon existing 3D editing frameworks, Catalyst4D supports a broad spectrum of dynamic scene editing scenarios. 
Our approach generalizes from monocular to multi-view inputs, and covers both localized dynamic object edits such as character appearance, clothing changes, and fine-grained color adjustments, as well as global transformations including scene-level style transfer and artistic effects. 
Across all cases, Catalyst4D produces temporally coherent, spatially consistent, and visually compelling 4D results.

\noindent\textbf{Qualitative comparison.}
The qualitative results achieved on different datasets are shown in Fig.~\ref{fig:qualitative}, with the corresponding text prompts displayed on the leftmost side. 
As can be seen, Instruct 4D-to-4D~\cite{mou2024instruct} and Instruct-4DGS~\cite{kwon2025efficient} lack fine-grained localization capabilities, often causing unintended modifications to non-target regions. Although CTRL-D~\cite{he2025ctrl} supports localized editing in dynamic scenes, its reliance on 2D image edits introduces limitations. When the 2D edits deviate significantly from the original scene content, the reconstructed 4D representation fails to accurately reflect them, resulting in degraded visual fidelity (see the 3rd and 5th rows of CTRL-D in Fig.\ref{fig:qualitative}).
In contrast, our 3D editing component employs iterative alignment of 2D and 3D representations, producing more natural and coherent edits. Building on these geometry-consistent 3D edits, \textbf{Catalyst4D} focuses on propagating them to 4D Gaussian scenes, achieving enhanced spatio-temporal consistency and visual realism.

\subsection{Quantitative Results}
Tab.~\ref{tab:quantitative} presents quantitative comparisons with Instruct 4D-to-4D (IN4D), Instruct-4DGS (I4DGS), and CTRL-D on two DyNeRF scenes (Sear-steak, Coffee-martini) and one MeetRoom scene (Trimming). Across all three scenes, our method achieves the highest CLIP similarity, indicating the strongest semantic alignment with the editing instructions. Regarding efficiency, our method outperforms CTRL-D and Instruct 4D-to-4D, and maintains a highly competitive training time comparable to Instruct-4DGS, while delivering vastly superior semantic fidelity. In terms of temporal coherence, measured by \textbf{Consistency} from VBench~\cite{huang2024vbench}, our method also delivers consistently strong performance, matching or surpassing CTRL-D in most cases. We observe a slight drop in consistency on only one scene, which stems from our design choice of not retraining the deformation network or modifying the density of edited dynamic Gaussians. Despite this, the overall results demonstrate that our method maintains competitive stability while offering significantly better semantic fidelity.

\begin{table}[t!]
\caption{
Quantitative comparison with Instruct 4D-to-4D (IN4D), Instruct-4DGS (I4DGS) and CTRL-D. 
\textbf{Bold} and \underline{underlined} denote the best and second-best results, respectively. * indicates 2 GPUs are used. Our method achieves superior semantic fidelity while maintaining highly competitive temporal stability.
}
\label{tab:quantitative}
\centering
\setlength{\tabcolsep}{1pt}
\begin{tabular}{ccccc}

\toprule

Scene & Method & CLIP sim.$\uparrow$ & Consistency$\uparrow$ & Time $\downarrow$\\ 

\midrule 

\multirow{4}{*}{Sear-steak} 
& Ours & \textbf{0.252} & \underline{0.983} & \underline{50 min}\\
& IN4D & 0.246 & 0.962 & 2 h$^*$\\
& I4DGS & 0.220 & 0.980 & \textbf{40 min}\\
& CTRL-D & \underline{0.249} & \textbf{0.985} & 55 min\\
\midrule

\multirow{4}{*}{Coffee-martini} 
& Ours & \textbf{0.249} & \textbf{0.986} & \underline{50 min}\\
& IN4D & 0.241 & 0.981 & 2 h$^*$\\
& I4DGS & 0.244 & 0.977 & \textbf{40 min}\\
& CTRL-D & \underline{0.246} & \underline{0.983} & 55 min\\
\midrule

\multirow{4}{*}{Trimming}
& Ours & \textbf{0.251} & \textbf{0.967} & \underline{40 min}\\
& IN4D & 0.243 & 0.945 & 2 h$^*$\\
& I4DGS & 0.245 & \underline{0.964} & \textbf{30 min}\\
& CTRL-D & \underline{0.248} & 0.962 & 50 min\\

\bottomrule
\end{tabular}
\end{table}

\begin{figure}[t]
\centering
\includegraphics[width=\columnwidth]{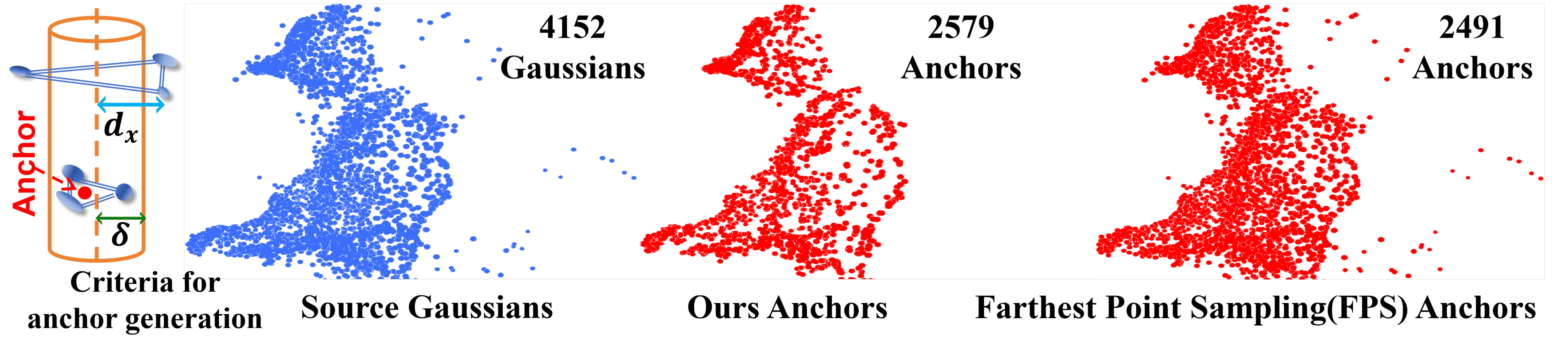}
\caption{Comparison of anchor construction methods.}
\label{fig:amg_visual}
\end{figure}

\begin{figure}[t]    
\centering
\includegraphics[width=\columnwidth]{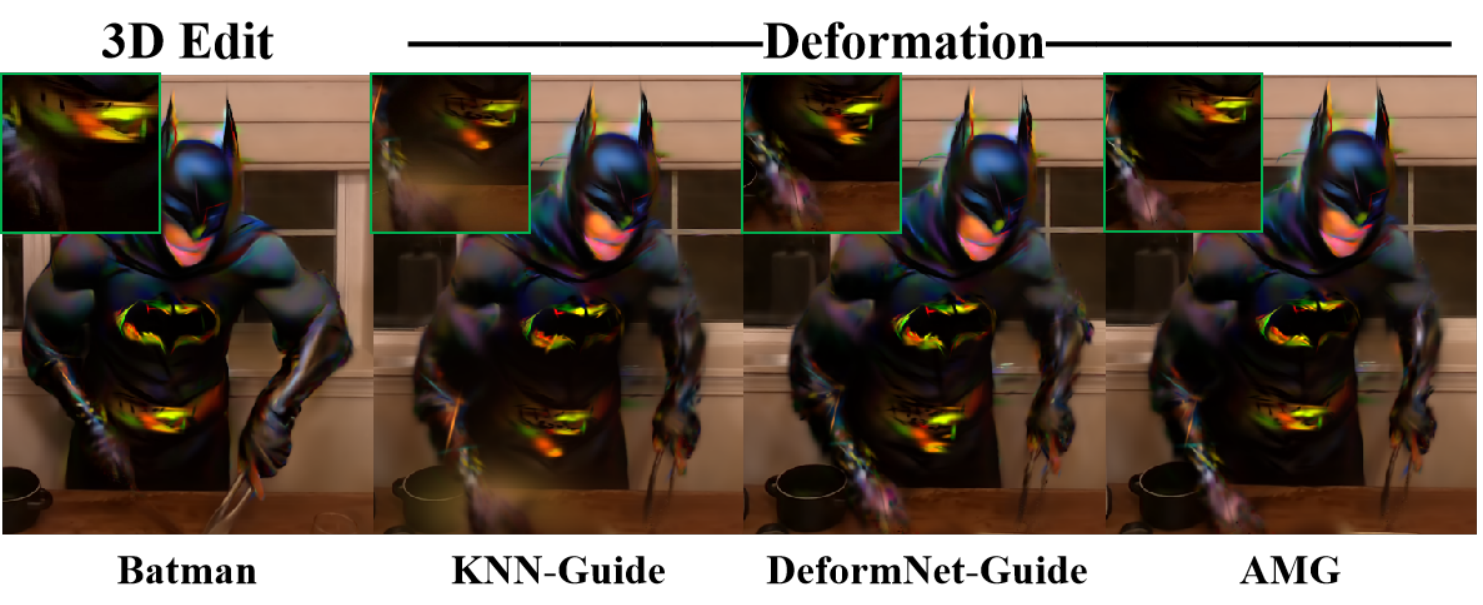}
\caption{Ablation study on the Anchor-based Motion Guidance. Our method effectively avoids geometric distortions caused by incorrect motion propagation.}
\label{fig:ablation1}
\end{figure}

\begin{figure}[t]    
\centering
\includegraphics[width=\columnwidth]{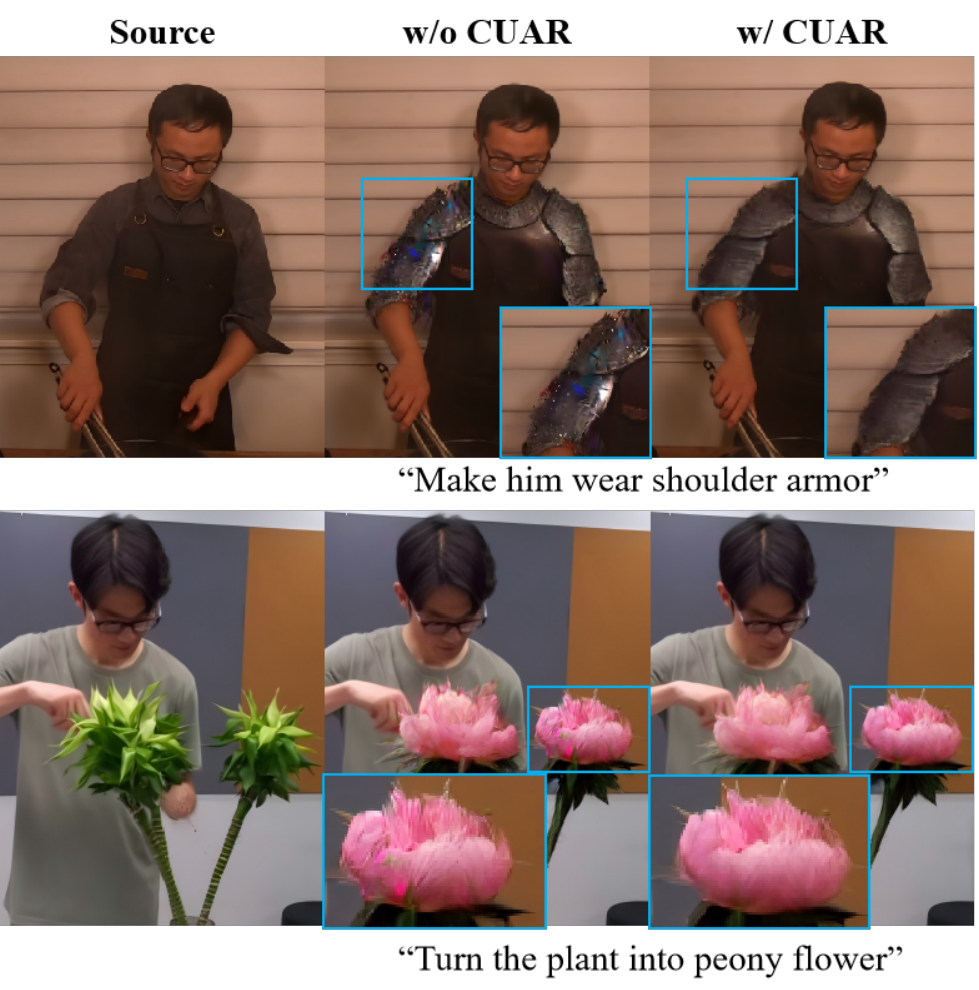}
\caption{Ablation study on the Color Uncertainty-guided Appearance Refinement (CUAR). It significantly enhances visual fidelity by mitigating color artifacts.}
\label{fig:ablation2}
\end{figure}

\subsection{Ablation Studies}
\noindent\textbf{On Anchor Construction.}
An anchor is generated only when multiple nearby Gaussians consistently fall within an adaptively defined cylinder (\cref{fig:amg_visual}). This design yields more reliable anchors for temporal propagation than naive sampling (e.g., FPS).

\noindent\textbf{On Anchor-based Motion Guidance.}
We assess the effectiveness of AMG module in propagating deformations to edited dynamic Gaussians. As a baseline, we first apply a KNN-based strategy (\textbf{KNN-Guide}), where the deformation for each edited Gaussian is calculated as a weighted average of the deformations from its $k=3$ nearest neighbors in the original dynamic point cloud. This method leads to motion entanglement across different semantic parts of the object due to the presence of spatially close but semantically unrelated Gaussians, especially in noisy or sparse regions. As shown in Fig.~\ref{fig:ablation1}, motion signals from the hand region erroneously influence unrelated areas like the torso, resulting in unnatural deformations.
We also compare against directly applying the deformation network from 4D representations to edited Gaussians (\textbf{DeformNet-Guide}). This avoids some local noise effects, but because the edited Gaussians deviate from the canonical Gaussians used during deformation training, large misalignments and geometric artifacts still occur, especially in areas with structural edits.

In contrast, AMG module establishes robust structural correspondences between edited and original Gaussians by matching sparse yet representative anchor points. This preserves regional motion and prevents interference from sparse noise, leading to improved geometric fidelity. The quantitative results in Tab.~\ref{tab:ablation} confirm this observation, our method achieves higher CLIP similarity and better temporal consistency than relying on the deformation network (\textbf{w/o AMG}) to propagate the edited Gaussians.

\begin{table}[t!]
\caption{Quantitative Ablation studies on AMG and CUAR modules. The results show that AMG is critical for establishing correct motion propagation, while CUAR further enhances both semantic and temporal scores by refining appearance consistency.
}
\label{tab:ablation}
\centering
\setlength{\tabcolsep}{1pt}
\begin{tabular}{ccccc}
\toprule
Method  & CLIP Sim. $\uparrow$  & Consistency $\uparrow$  \\ 
\midrule 
w/o AMG & 0.245 & 0.966 \\ 
w/o CUAR & \underline{0.248} & \underline{0.969}\\
Full model & \textbf{0.252} & \textbf{0.971}\\
\bottomrule
\end{tabular}
\end{table}

\noindent\textbf{On Color Uncertainty-guided Appearance Refinement.}
We further ablate the impact of our CUAR module. As shown in Fig.~\ref{fig:ablation2}, removing the refinement step (\textbf{w/o CUAR}) results in visible flickering and inconsistent color artifacts in dynamic regions. For instance, in the middle column of Fig.~\ref{fig:ablation2}, the surfaces of the armor and flower exhibit anomalous colors. This is because editing operations inevitably affect internal Gaussians that only become visible during motion.
Our CUAR module (\textbf{w/ CUAR}) identifies artifact-prone regions using Gaussian-wise color uncertainty derived from multi-view SH color discrepancies. These regions are selectively refined using warping supervision from the more reliable first-frame rendered images, effectively suppressing artifacts and ensuring consistency. The quantitative results presented in Tab.~\ref{tab:ablation} corroborate this visual improvement, confirming its effectiveness.
\section{Conclusion}
We present Catalyst4D, a novel framework that brings the editing capabilities of 3D methods into the realm of dynamic 4D Gaussian scenes. By introducing an anchor-based motion guidance strategy to preserve structural consistency and a color uncertainty-guided appearance refinement technique to suppress motion-induced artifacts, Catalyst4D enables high-quality edits across time. Extensive experiments confirm its superior visual fidelity over existing approaches.

\section*{Acknowledgment}
This work is partly supported by the National Key Research and Development Plan (2024YFB3309302), Xiaomi Young Talents Program, the Research Program of State Key Laboratory of Complex and Critical Software Environment, and the Fundamental Research Funds for the Central Universities.

{
    \small
    \bibliographystyle{ieeenat_fullname}
    \bibliography{main}
}
\clearpage
\setcounter{page}{1}
\renewcommand\thesection{A\arabic{section}}
\renewcommand*{\thefigure}{A\arabic{figure}}
\renewcommand*{\thetable}{A\arabic{table}}

\setcounter{section}{0}
\setcounter{figure}{0}
\setcounter{table}{0}

\maketitlesupplementary
\section{Video Demo}
We provide demo videos on our project page showing the dynamic editing capabilities of our method. The videos include comparative results against baseline approaches on novel test viewpoints, as well as demonstrations of image-guided global style transfer and text-guided local edits enabled by the proposed Catalyst4D. We encourage readers to view these videos, as they offer a clear illustration of the high visual fidelity and editing consistency achieved by our approach.

\section{Additional Comparison Results}
\subsection{Additional Quantitative Comparison}
Standard CLIP and VBench scores may fail to capture fine-grained 3D-to-4D propagation due to their global nature, we further adopt \textbf{EditScore} (instruction-following accuracy and visual fidelity)~\cite{luo2025editscore} and \textbf{VE-Bench} (local temporal consistency)~\cite{sun2025ve}, better aligned with human perception.
Tab.~\ref{tab:editscore_cpr} on \textbf{16} prompts(multi-camera) from the main paper shows Catalyst4D consistently surpasses prior methods.

\subsection{Additional Quantitative Ablation}
We clarify that the main paper ablation (Tab.~2) is cumulative: w/o AMG is the ``w/o both'' baseline, and w/o CUAR retains only AMG.  
Tab.~\ref{tab:ablation_all}, evaluated with new metrics,
further confirms both AMG and CUAR are essential, outperforming DeformNet-Guide (w/o both).

\subsection{Comparison on Background Preservation}
\begin{figure}[t]
\centering
\includegraphics[width=\columnwidth]{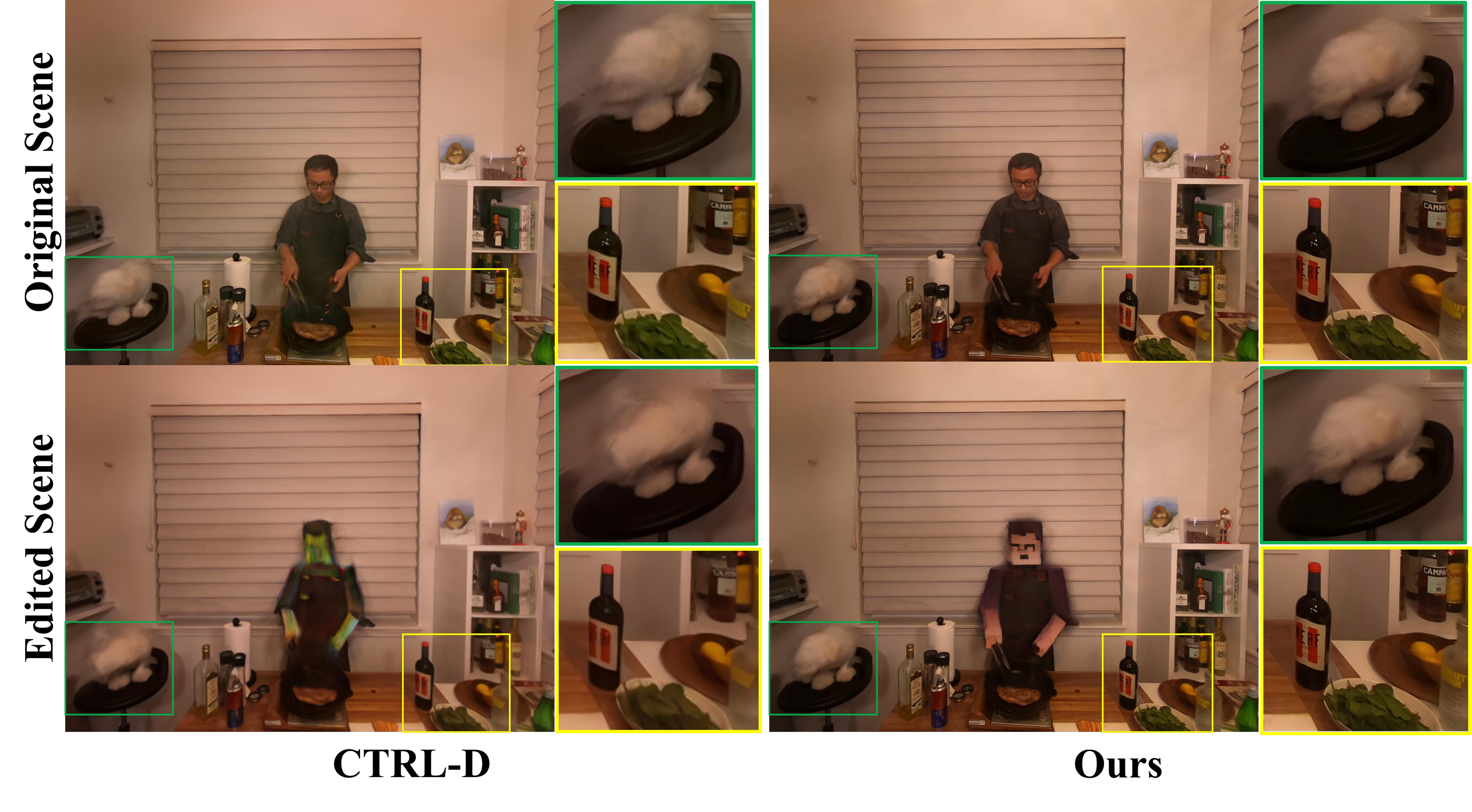}
\caption{Qualitative comparison of localized editing. In contrast to CTRL-D, which introduces inconsistencies in non-edited regions, our method achieves more precise and localized editing by constraining dynamic Gaussians via 3D editing gradients.}
\label{fig:background}
\end{figure}

Although the compared CTRL-D \cite{he2025ctrl} appears visually plausible, it still introduces undesired modifications in non-edited regions due to the limitations of the underlying 2D diffusion model. As illustrated in Fig.\ref{fig:background}, while CTRL-D edits the character, it also causes noticeable deviations in non-target objects such as the dog on the stool and objects on the table compared to the original scene. By contrast, our method constrains target dynamic Gaussians directly through 3D editing gradients, enabling more precise and localized editing in dynamic scenes.

\subsection{Comparison on the Monocular Dataset}
\begin{figure*}[t]
\centering
\includegraphics[width=0.7\textwidth]{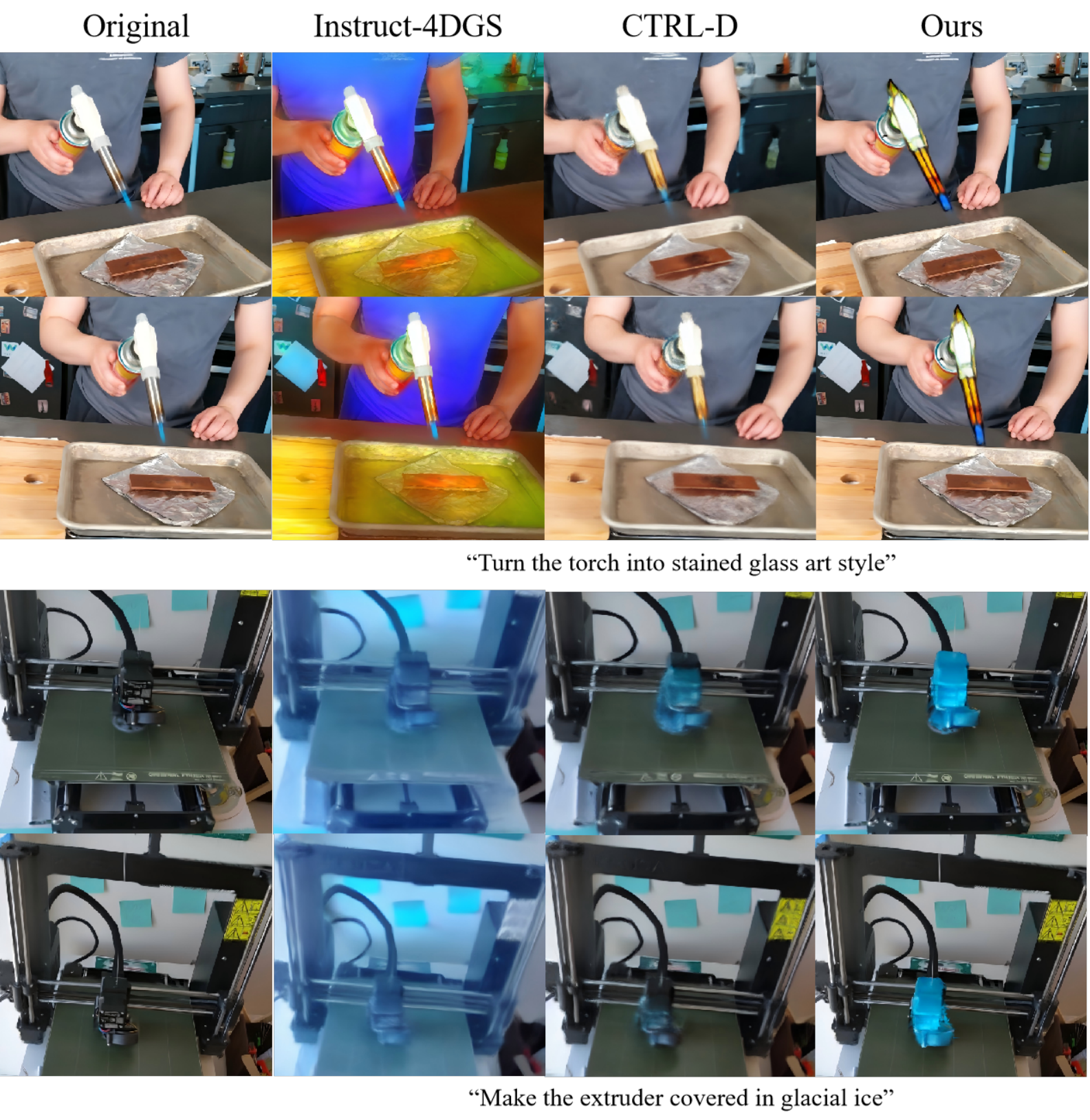}
\caption{Comparison on HyperNeRF Dataset. The results show that Instruct-4DGS fails to correctly localize edits and CTRL-D produces blurry and temporally inconsistent results, whereas our method demonstrates significant advantages in both visual quality and temporal consistency.}
\label{fig:mono_compare}
\end{figure*}

We further evaluate text-driven 4D editing on the monocular HyperNeRF \cite{park2021hypernerf} dataset. Since Instruct 4D-to-4D does not support this dataset, we primarily compare our method against Instruct-4DGS and CTRL-D. The qualitative comparisons are shown in Fig.~\ref{fig:mono_compare}, where the top and bottom rows of each example correspond to different timesteps.
Instruct-4DGS struggles to localize the target object, resulting in edits being incorrectly applied to irrelevant regions of the scene. CTRL-D, which reconstructs the 4D representation from 2D diffusion outputs, suffers from the inherent modality gap, producing blurry results and noticeable temporal inconsistency. As illustrated in the “Glacial extruder’’ example in the lower half of Fig.~\ref{fig:mono_compare}, the appearance of the edited extruder varies substantially across timesteps.
In contrast, our method directly propagates a high-fidelity 3D edit from the first frame to all subsequent frames, yielding clearer textures, stronger temporal coherence, and significantly improved overall 4D editing quality.

\begin{table}[t!]
\caption{Quantitative comparison using EditScore and VE-Bench.}
\label{tab:editscore_cpr}
\centering
\setlength{\tabcolsep}{10pt} 
\resizebox{\columnwidth}{!}{
\begin{tabular}{c|c|c|c|c}
\toprule
Metric & IN4D & I4DGS & CTRL-D & Ours  \\ 
\hline 
EditScore$\uparrow$ & 4.034 & 5.618 & 4.326 & \textbf{7.375} \\ 
\hline
VE-Bench$\uparrow$ & 0.155 & 0.256 & 0.163 & \textbf{1.080}\\
\bottomrule
\end{tabular}
}
\end{table}

\begin{table}[t!]
\caption{Ablation studies on AMG and CUAR modules.}
\label{tab:ablation_all}
\centering
\setlength{\tabcolsep}{9pt}
\resizebox{\columnwidth}{!}{\begin{tabular}{c|c|c|c|c}
\toprule
Metric & w/o both & w/ AMG & w/ CUAR & w/ both \\ 
\hline 
EditScore$\uparrow$ & 2.713 & 5.112 & 5.005 & \textbf{7.375} \\ 
\hline
VE-Bench$\uparrow$ & 0.059 & 0.452 & 0.580 & \textbf{1.080}\\
\bottomrule
\end{tabular}
}
\end{table}

\begin{figure*}[t]
\centering
\includegraphics[width=0.8\textwidth]{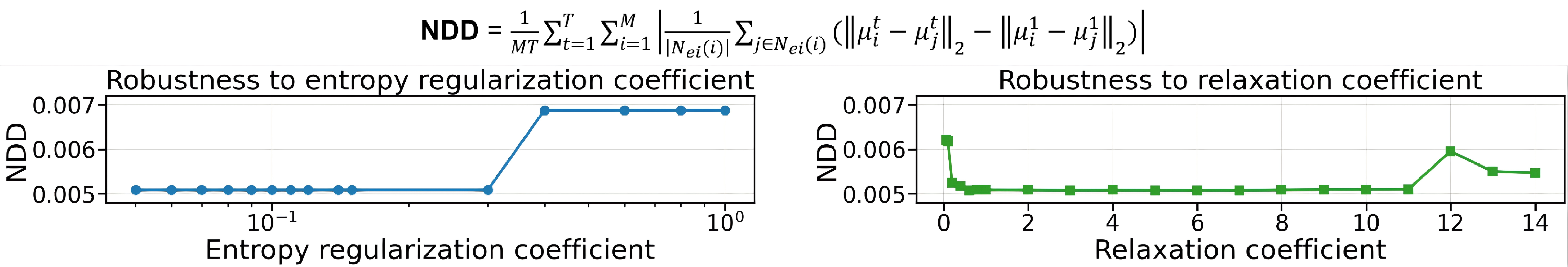}
\caption{Effect of Sinkhorn regularization on motion transfer.}
\label{fig:uot_ablation}
\end{figure*}

\begin{figure*}[t]
\centering
\includegraphics[width=\textwidth]{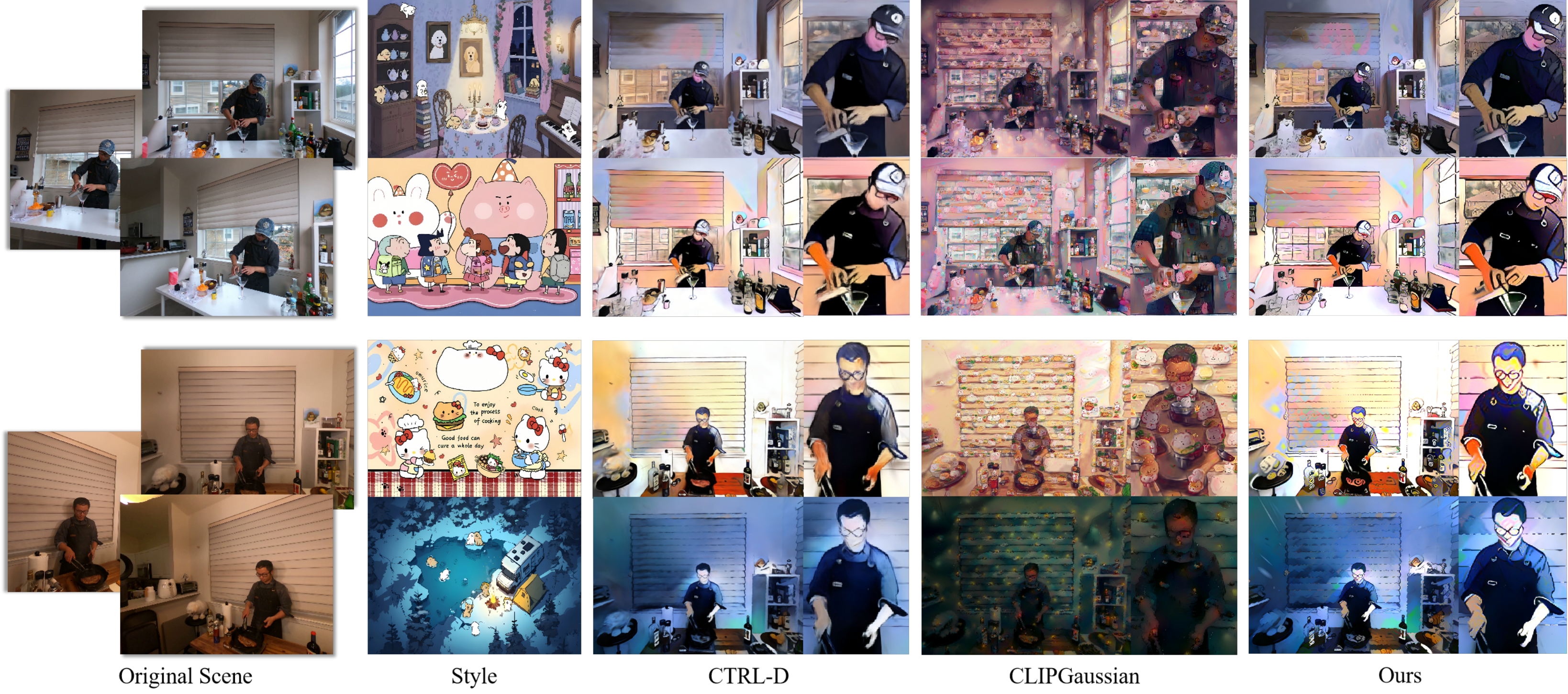}
\caption{Comparative results for global style transfer. Unlike the over-smoothed results of CTRL-D and the chaotic textures from CLIPGaussian, our method produces finer visual textures, better preserves the original scene's geometry, and demonstrates a color distribution more consistent with the reference style image.}
\label{fig:style_compare}
\end{figure*}

\subsection{Robustness of Sinkhorn Algorithm}
Our outlier-filtered anchors yield a well-conditioned cost matrix, enabling stable Sinkhorn convergence. To assess robustness under extreme topology changes (e.g. \textit{Sear-steak} to \textit{Minecraft}), 
we measure Neighborhood Distance Deviation (NDD), which captures the temporal preservation of local anchor geometry and directly reflects correspondence stability.
Fig.~\ref{fig:uot_ablation} confirms NDD stability across wide Sinkhorn ranges ($\lambda_0 \in [0.05,0.3], \lambda_{1,2} \in [0.6,11]$).

\subsection{Comparison of Global Style Transfer}
We further compare our method with CLIPGaussian \cite{howil2025clipgaussian} and CTRL-D on the task of image-guided global style transfer. To establish a controlled comparison with CTRL-D, which requires an image prompt, we first stylize the first frame 3D scene using SGSST \cite{galerne2025sgsst}, our 3D baseline. A rendered image from this stylized 3D scene is then provided to CTRL-D as its style reference. Consequently, our method and CTRL-D begin with nearly identical style information but leverage distinct technical approaches. Our method propagates the style from the 3D scene across the sequence (a 3D-to-4D approach), while CTRL-D uses its DreamBooth-based fine-tuning to reconstruct the 4D effect from the 2D style image (a 2D-to-4D approach). This allows for a direct assessment of the two pathways, minimizing the influence of external stylization models.

As shown in Fig.~\ref{fig:style_compare}, the global style transfer results produced by CTRL-D appear superficially similar to ours, an expected outcome given that CTRL-D's style reference is a rendered image from our 3D stylization baseline. However, our method achieves noticeably finer textures and higher visual clarity. CTRL-D, which relies on a 2D diffusion model, often generates results with blurriness and over-smoothing due to the inherent 2D-to-4D reconstruction gap. Meanwhile, CLIPGaussian tends to introduce chaotic textures, as reflected in the distorted facial region of the character in the example shown in the fourth column of Fig.~\ref{fig:style_compare}.
In contrast, our approach preserves the underlying 4D geometry more effectively by propagating SGSST \cite{galerne2025sgsst}-based stylization (our 3D baseline) consistently across the sequence. Additionally, our results exhibit a color distribution that more faithfully matches the reference style image.

\section{Implementation Details}
\subsection{Optimal Transport-based Anchor Matching}
Given the anchor set $A_{\mathrm{src}}=\{\mathbf{p}_i^{\mathrm{src}}\}_{i=1}^{n}$ of the original dynamic Gaussians $\mathcal{G}^1$ in the first frame and the anchor set $A_{\mathrm{edit}}=\{\mathbf{p}_j^{\mathrm{edit}}\}_{j=1}^{m}$ of the edited Gaussians $\mathcal{G}^1_{\mathrm{edit}}$, we formulate the anchor correspondence as an unbalanced optimal transport (UOT) problem and solve it using the Sinkhorn algorithm \cite{cuturi2013sinkhorn}. 

We define a distance matrix $D$ using a Welsch robust penalty:
\begin{equation}
    D_{ij} = 1 - \exp\left(-\frac{\|\mathbf{p}_i^\mathrm{src} - \mathbf{p}_j^\mathrm{edit}\|_2^2}{2\beta^2}\right),
\end{equation}
where $\beta = \gamma \cdot d_{\mathrm{med}}$, $d_{\mathrm{med}}$ is the median of all pairwise anchor distances, and we empirically set $\gamma = 0.05$.

The UOT objective is written as:
\begin{multline}
    \min_{P \in \mathbb{R}_+^{n \times m}}  
    \sum_{i=1}^{n} \sum_{j=1}^{m} D_{ij} \mathbf{P}_{ij}
    - \lambda_0 \sum_{i,j} \mathbf{P}_{ij} \log \mathbf{P}_{ij} \\
    + \lambda_1 \operatorname{KL}\!\left( \mathbf{P} \mathbf{1}_m \,\Big\|\, \frac{1}{n}\mathbf{1}_n \right)
    + \lambda_2 \operatorname{KL}\!\left( \mathbf{P}^{\mathrm{T}} \mathbf{1}_n \,\Big\|\, \frac{1}{m}\mathbf{1}_m \right),
\end{multline}
\text{s.t.} $\mathbf{P}_{ij} \ge 0, \forall i \in \{1, ..., n\}, j \in \{1, ..., m\}$, where $\mathbf{P}$ is the transport plan matrix, $\lambda_0 = 0.1$, $\lambda_1 = 1.0$, $\lambda_2 = 1.0$ are regularization weights, and $\mathit{KL}$ denotes the Kullback-Leibler divergence.

From the optimal plan $\mathbf{P}^*$, we obtain the correspondence $corr$ by assigning each edited anchor to the source anchor with the highest transport mass:
\begin{equation}
    corr_j = \arg\max_{i \in \{1, \ldots, n\}} \mathbf{P}^*_{ij}.
\end{equation}

\subsection{Deformation of Edited Dynamic Gaussians}
Notably, when computing the deformation of edited Gaussians, unlike the additive form used for positional deformation $\Delta{\mu}$ and rotational deformation $\Delta\mathbf{q}$, scaling deformation $\Delta\mathbf{s}$ is defined multiplicatively to preserve proportional change:

\begin{equation}
    \begin{aligned}
        {{\mu}}^t &= {{\mu}}^1 + \Delta{{\mu}}^t, \\
        \mathbf{q}^t &= \mathbf{q}^1 + \Delta\mathbf{q}^t, \\
        \mathbf{s}^t &= \mathbf{s}^1 \cdot \Delta\mathbf{s}^t.
    \end{aligned}
\end{equation}
where $\Delta{{\mu}}^t$, $\Delta\mathbf{q}^t$, $\Delta\mathbf{s}^t$ are the deformation quantities of edited Gaussians from frame 1 to frame $t$.

 



\subsection{Additional Experimental Details}

\noindent\textbf{Anchor Construction.}
Since $\mathcal{G}_{\mathrm{edit}}^1$ and $\mathcal{G}^1$ are largely spatially aligned, we compute a shared bounding sphere and sample an identical set of rays to construct anchor points for both Gaussian clouds. Specifically, we sample 300{,}000 rays from the minimum bounding sphere and use $k=2$ nearest neighbors when forming local anchor neighborhoods. For Gaussians not covered by any anchor region, we assign them to the nearest anchor using Euclidean distance.

\medskip
\noindent\textbf{Loss Design.}
For foreground edits, we use standard L1+SSIM to propagate the effect faithfully. Background supervision relies on L1 only, as CUAR freezes non-edited Gaussians and L1 sufficiently prevents artifacts; adding SSIM yields negligible gains while increasing cost.

\medskip
\noindent\textbf{Appearance Refinement Parameters.}
Eq.~17 exploits uncertainty contrast between artifact-prone and stable regions. The threshold $\epsilon$ trades quality for efficiency (higher for faster refinement, lower for broader coverage) and remains stable over a reasonable range. To flexibly control the spatial coverage of the artifact mask, we set the threshold parameter $\epsilon$ within the range $[0.5,\,2.0]$. During refinement, the loss weights are fixed as $\eta = 0.2$ and $\zeta = 0.3$, applied to the foreground and background objectives described in Sec.~4.2. These settings balance artifact correction with preservation of non-corrupted regions.

\medskip
\noindent\textbf{Evaluation Protocol and Implementation.}
For quantitative evaluation, we render novel test viewpoints and compute the CLIP similarity by averaging scores over 30 uniformly sampled frames from each test video. Temporal consistency is evaluated on the full rendered sequence to measure stability across viewpoints and timesteps.

For dynamic scene reconstruction with Swift4D, we adopt effective rank regularization~\cite{hyung2024effective} to suppress needle-like artifacts. Following the settings of 4DGS \cite{wu20244d}, we constrain the edited dynamic Gaussians to share consistent opacity and color attributes across all frames during optimization.

The training breakdown is as follows: Anchor construction ($<$30s), Sinkhorn solver ($\sim$15s), Motion Guidance ($\sim$1 min), and CUAR (25–35 min). All experiments are implemented in Python~3.7.12 using PyTorch~1.13.1 on Ubuntu~22.04. Our method is trained on a single NVIDIA A100 Tensor Core GPU.

\subsection{Editing Prompts and 3D Baseline Configurations}

Here, we provide the detailed editing configurations used in our experiments, including the text prompts and the corresponding 3D editing baselines adopted for each edit.

~

For the results shown in Fig.~1, the editing instructions and 3D baseline methods are:
\begin{itemize}
    \item ``Turn the torch into Pop art style'' with DreamCatalyst,
    \item ``Make the torch carved from a flawless emerald'' with DGE,
    \item ``Turn him into a Claymation character'' with DGE,
    \item ``Make the extruder covered in glacial ice'' with DreamCatalyst,
    \item ``Make the extruder look like toast'' with DGE,
    \item ``Turn him into JoJo's Bizarre Adventure anime style'' with DGE.
\end{itemize}

For the edits in Fig.~4, we use:
\begin{itemize}
    \item ``Turn his clothes into a football player outfit'' with DreamCatalyst,
    \item ``Turn his hat into a newsboy cap'' with DGE,
    \item ``Turn him into a Minecraft character'' with DGE,
    \item ``Make them look like Marble roman sculptures'' with DreamCatalyst,
    \item ``Turn him into Deadpool'' with DGE.
\end{itemize}

~

For the results of our method reported in Tab.~1, all 3D edits are performed using DGE.  
The applied prompts are:
\begin{itemize}
    \item \textbf{Sear-steak.} ``Turn him into a Minecraft character'', ``Turn him into Super Mario'',
    \item \textbf{Coffee-martini.} ``Turn his hat into a newsboy cap'', ``Turn him into JoJo's Bizarre Adventure anime style'',
    \item \textbf{Trimming.} ``Change the leaves to autumn leaves'', ``Turn him into Deadpool''.
\end{itemize}

~

For the results in Tab.~2, the editing setups are:
\begin{itemize}
    \item ``Turn him into a Minecraft character'' with DGE,
    \item ``Turn him into Batman'' with DreamCatalyst,
    \item ``Turn him into JoJo's Bizarre Adventure anime style'' with DGE.
\end{itemize}





\section{Limitation}
While Catalyst4D effectively extends high-quality 3D editing capabilities to dynamic 4D scenes, it inherits certain dependencies from the underlying components or pretrained models.

\begin{figure}[t]
  \centering
\includegraphics[width=\linewidth]{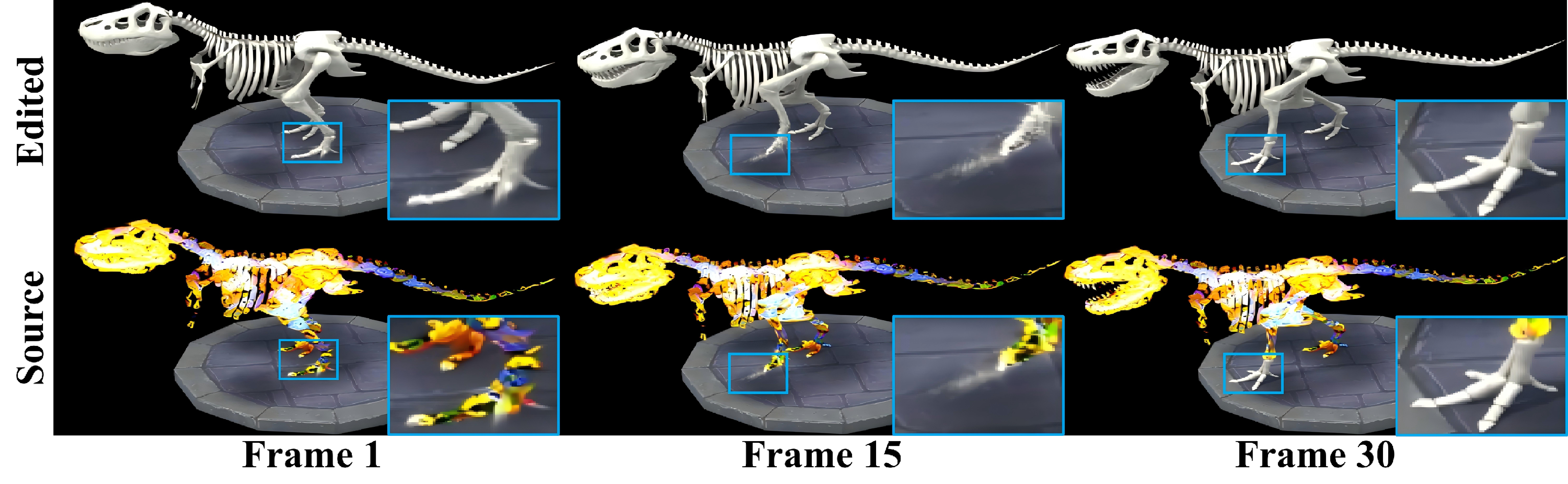}
   \caption{Failure case (motion jitter illustration).}
   \label{fig:failure}
\end{figure}

First, the temporal coherence of our method is influenced by the quality of the initial 3D edits. If the 3D editing results lack sufficient spatial consistency, this may propagate and affect the spatio-temporal consistency in the final dynamic outputs. Moreover, as our method does not modify the deformation network or re-optimize Gaussian densities, it relies on the stability of the underlying 4D reconstruction. Under severe reconstruction noise (e.g., point jitter or low-opacity Gaussians), motion guidance may be locally disrupted. In \cref{fig:failure} (\textit{D-NeRF, trex}), background Gaussians drift into the edited foreground, causing local artifacts, reflecting a shared limitation of current 4D reconstruction models.
Nonetheless, such limitations are not fundamental and can be mitigated as upstream 3D editing and 4D reconstruction techniques continue to advance. Our framework is fully compatible with future improvements in both areas.

\end{document}